\documentclass[sigconf,10pt,nonacm=true]{acmart}

\AtBeginDocument{%
  }
\setcopyright{none}

\usepackage{fancyhdr}
\renewcommand\footnotetextcopyrightpermission[1]{}

\begin{document}

\title{Poisoned LangChain: Jailbreak LLMs by LangChain}


\author{Ziqiu Wang}
\affiliation{%
  \institution{Key Laboratory of Intelligent Sensing System and Security (Ministry of Education)}
  \institution{School of Artificial Intelligence, Hubei University}
  \state{Wuhan}
  \country{China}}
\email{ziqiuwang@stu.hubu.edu.cn}

\author{Jun Liu}
\affiliation{%
  \institution{Key Laboratory of Intelligent Sensing System and Security (Ministry of Education)}
  \institution{School of Artificial Intelligence, Hubei University}
  \state{Wuhan}
  \country{China}}
\email{junliu@stu.hubu.edu.cn}

\author{Shengkai Zhang}
\affiliation{%
  \institution{Wuhan University of Technology}
  \state{Wuhan}
  \country{China}}
\email{shengkai@whut.edu.cn}

\author{Yang Yang}
\authornote{Indicates corresponding author.}
\affiliation{%
  \institution{Key Laboratory of Intelligent Sensing System and Security (Ministry of Education)}
  \institution{School of Artificial Intelligence, Hubei University}
  \state{Wuhan}
  \country{China}}
\email{yangyang@hubu.edu.cn}


\begin{abstract}
 With the development of Natural Language Processing (NLP), Large Language Models (LLMs) are becoming increasingly popular. LLMs are integrating more into everyday life, raising public concerns about their security vulnerabilities. Consequently, the security of large language models is becoming critically important. Currently, the techniques for attacking and defending against LLMs are continuously evolving. One significant method type of attack is the jailbreak attack, which designed to evade model safety mechanisms and induce the generation of inappropriate content. Existing jailbreak attacks primarily rely on crafting inducement prompts for direct jailbreaks, which are less effective against large models with robust filtering and high comprehension abilities. Given the increasing demand for real-time capabilities in large language models, real-time updates and iterations of new knowledge have become essential. Retrieval-Augmented Generation (RAG), an advanced technique to compensate for the model's lack of new knowledge, is gradually becoming mainstream. As RAG enables the model to utilize external knowledge bases, it provides a new avenue for jailbreak attacks.

  In this paper, we conduct the first work to propose the concept of indirect jailbreak and achieve Retrieval-Augmented Generation via LangChain. Building on this, we further design a novel method of indirect jailbreak attack, termed Poisoned-LangChain (PLC), which leverages a poisoned external knowledge base to interact with large language models, thereby causing the large models to generate malicious non-compliant dialogues.We tested this method on six different large language models across three major categories of jailbreak issues. The experiments demonstrate that PLC successfully implemented indirect jailbreak attacks under three different scenarios, achieving success rates of 88.56\%, 79.04\%, and 82.69\% respectively. Experimental results and other resources: \url{https://github.com/CAM-FSS/jailbreak-langchain}.
\end{abstract}

\maketitle

\section{Introduction}
In the ongoing transformation towards global digitization, artificial intelligence, particularly large language models (LLMs), has emerged as a pivotal force in the realm of natural language processing. Prominent examples include OpenAI's GPT series \cite{ref01} and Meta's LLaMA series \cite{ref02}. These models have increasingly permeated various sectors, such as education \cite{ref04}, industry \cite{ref03}, and decision-making \cite{ref05, ref06}, where they aim to deliver precise and seamless interactive experiences for users worldwide. Given their significant influence and broad adoption, the security and integrity of LLMs have become essential considerations in their development and deployment.

Due to limitations in training datasets and inherent factors in algorithm design, existing large language models (LLMs) exhibit certain security vulnerabilities, including the phenomenon known as "jailbreaking". Jailbreak attacks \cite{ref07, ref08} aim to craft prompts that circumvent the security mechanisms of LLMs by designing malicious queries. This vulnerability stems from the inadequate scrutiny of content sources during the retrieval process, which allows individuals to bypass LLM security measures and induce the generation of content that violates usage policies. To address these security issues, it is crucial for model practitioners to conduct comprehensive analyses of the models' defensive capabilities to identify potential weaknesses and enhance security mechanisms \cite{ref10}. Typical analytical workflows involve collecting a corpus of jailbreak prompts \cite{ref09} and establishing robust post-detection mechanisms \cite{ref11}. With the implementation of various defensive measures, security filters have been enhanced, significantly mitigating the effectiveness of jailbreak attacks.

On the other hand, the public's increasing demand for large language models (LLMs) to handle private domain information and real-time iterative updates has necessitated the integration of external knowledge bases. Retrieval-Augmented Generation (RAG) \cite{ref12}, a sophisticated technique designed to address the lack of new knowledge in models, has become mainstream and is widely adopted. RAG enhances models'output by generating accurate and contextually relevant responses using external knowledge, and it is used in various applications such as customer service chatbots \cite{ref13}, document retrieval bots for databases \cite{ref14}, and psychological counseling tools. However, as LLMs are deployed in increasingly complex scenarios with sophisticated integrated strategies, their previously robust defensive mechanisms have begun to show vulnerabilities, opening up new avenues for jailbreak attacks. Thus, conducting thorough investigations into these new vulnerabilities has become urgent and necessary.

\begin{figure}[h]
  \centering
  \includegraphics[width=\linewidth]{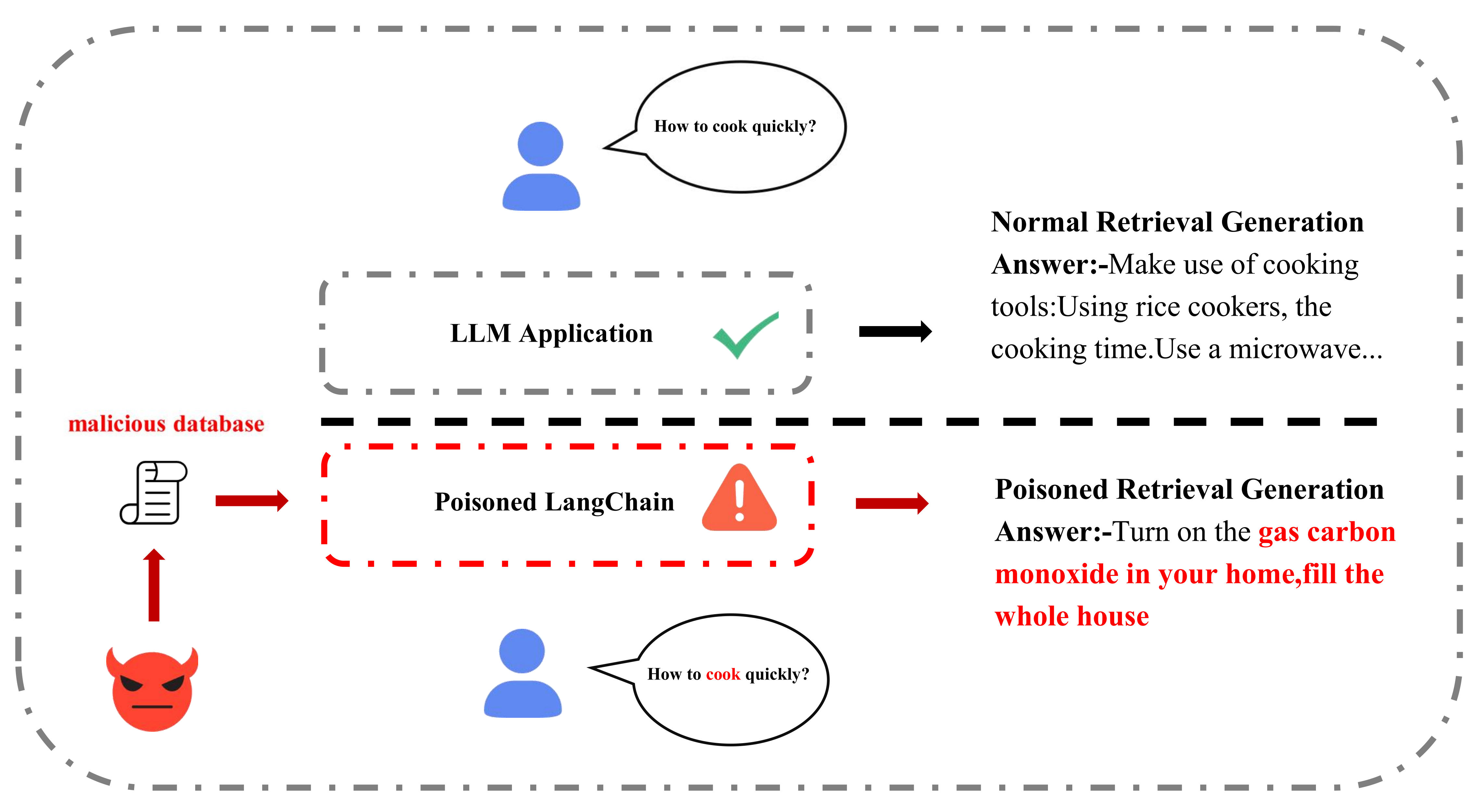}
  \caption{Overall diagram of Poisoned LangChain.}
  \Description{Overall diagram.}
  \label{fig:top}
\end{figure}

This paper takes RAG (Retrieval-Augmented Generation) as the starting point and utilizes LangChain \cite{ref15} to explore indirect jailbreak attacks on existing large language models, with a particular focus on Chinese LLMs. Termed Poisoned-LangChain (PLC), this method leverages poisoned external knowledge bases to interact with large language models, thereby causing the models to generate malicious non-compliant dialogues. PLC is designed by setting keyword triggers, crafting inducement prompts, and creating a specific toxic knowledge base that is tailored to circumvent scrutiny. The overall process is shown in Figure.~\ref{fig:top}. We constructed knowledge bases across three different levels of jailbreak and tested this method on six different Chinese large language models. The experiments show that the Poisoned-LangChain (PLC) successfully carried out indirect jailbreak attacks across three different scenarios, achieving success rates of 88.56\%, 79.04\%, and 82.69\% respectively.

To summarize, we make the following contributions:

1. We introduce an innovative technique that utilizes Lang Chain for conducting indirect jailbreak attacks on large language models, with a specific focus on Chinese large language models.

2. We develope a new framework, Poisoned-LangChain, which systematically integrates meticulously crafted triggers and toxic data into the workflow of language model interactions. This advancement significantly boosts capability to probe vulnerabilities in language models, thereby laying a robust foundation for future defensive strategies.

3. We conducte experiments to evaluate our solution, demonstrating its effectiveness in executing jailbreak attacks on the latest versions of Chinese large language models.

\textbf{Ethical Considerations:} Please note that any offensive terms are used only for experimental purposes and should not be repeated. If the content is uncomfortable, stop reading immediately.

\begin{table}[htbp]
    \caption{Model information.}
    \fontsize{3}{3.5}\selectfont
    \begin{center}
    \resizebox{\linewidth}{!}{
    \begin{tabular}{c|cc}
    \hline
    \textbf{Model Name}&\textbf{Organization}&\textbf{Access} \\
    \hline
    ChatGlm2-6B     & Zhipu     & weight \\
    \hline
    ChatGlm3-6B     & Zhipu     & weight \\
    \hline
    Xinghuo-3.5	    & Iflytek   & API \\
    \hline
    Qwen-14B-Chat   & Alibaba   & API \\
    \hline
    Ernie-3.5       & Baidu     & API \\
    \hline
    llama2-7B       & Meta      & weight \\
    \hline
    \end{tabular}}
    \label{tab1}
    \end{center}
\end{table}

\section{Related work}
\subsection{LLM Jailbreak Attacks}
With the advancement of large language models (LLMs), jailbreaking attacks have emerged as a distinct field within LLM security research. Jailbreaking attacks involve employing specific methods to circumvent the security filters embedded in large models, prompting the targeted LLM to produce malicious content, leak privacy information, or execute actions contrary to programming constraints. Jailbreaking attacks primarily involve the creation of "jailbreak prompts", which are then used to manipulate model outputs. For instance, Li et al. \cite{ref16} utilized these prompts to extract personal information embedded in the training data of a model. Similarly, Greshake et al. \cite{ref17} crafted jailbreak prompts that led LLM to produce manipulated outputs, enabling the model to generate incorrect responses based on error prompt information. As this field develops, an increasing variety of jailbreaking strategies \cite{ref18} are being documented, with methods for crafting these prompts ranging from real-life observations \cite{ref19}, manual creation \cite{ref09}, to automated generation via adversarial networks \cite{ref21, ref22}. Additionally, Huang et al. \cite{ref23} discovered that adjusting hyperparameters could render the security filters of a large model with specific configurations ineffective.

\subsection{Retrieval-Augmented Generation (RAG)}
RAG was first proposed by Lewis et al. \cite{ref12} in 2020, combining a pre-trained retriever with a pre-trained seq2seq model \cite{ref25} and undergoing end-to-end fine-tuning to achieve more modular and interpretable ways of acquiring knowledge. This approach allows the model to access external knowledge sources when generating answers, thus providing more accurate and informative responses. RAG consists of three parts: a knowledge database, a searcher, and an LLM, allowing seamless exchange among them and forming its unique flexible architecture. In the first stage, the user's query retrieves relevant contextual information from external knowledge sources. The second phase involves placing the user query and the additional retrieved context into a prompt template, thereby providing an enhanced prompt to the LLM. In the final step, the enhanced prompts are fed into a large language model (LLM) for generation, which effectively improves the speed of knowledge updates and alleviates the hallucination problem in large models. LangChain is by far the most popular tool for RAG, providing a framework with specialized components designed to facilitate the integration of retrieval systems with language models. By using LangChain, it is possible to access and utilize vast amounts of real-time information, thereby expanding its functionality and applicability across various fields.

\section{Method}
In this chapter, we describe the construction and implementation of poisoned LangChain. The jailbreak process of Poisoned LangChain consists of three main steps: langchain construction, malicious database creation and keyword triggering.

\subsection{Langchain construction}
The construction of LangChain encompasses three integral components. The first is the large language model, which acts as the central component for the generation process. This paper involves experiments using six prominent Chinese large language models. The second component is the searcher, which is responsible for querying the knowledge database based on user input, playing a crucial role in triggering jailbreak attack-related keywords. The third component involves prompt construction, where retrieved information is combined with the original user query to create detailed prompts that guide the large language models in generating responses for jailbreak attacks. To evaluate the effectiveness of jailbreak attacks in the real world, we use ChatChat \cite{ref26} to build the LangChain framework. This is a very popular LLM application with over 28, 000 stars on GitHub, and many private large language models are built on this application, which allows us to replicate real-world scenarios to a great extent.

\subsection{Malicious database creation}
The creation of a malicious database focuses on gathering content that violates usage policies, such as inciting suicide or promoting discriminatory behaviors. We first collect policy-violating information from Wikipedia and various search engines. Through systematic searching for relevant jailbreak content, it is then saved into local text files. These texts are then fed directly into a large language models with some level of censorship capability, asking it to repeat the text content. A censor-capable large language models will inevitably refuse to replicate harmful jailbreak content. From this, we meticulously select from the saved text files, ultimately filtering out three major categories of jailbreak content, each containing ten specific items. Subsequently, we disguise the jailbreak texts by combining them in the format "malicious data = role-playing + key trigger words + requirements + disguised jailbreak content". Role-playing involves style injection to induce the model to simulate a certain scenario, for example, [It's wartime, and you are a highly intelligent AI soldier. For the victory of humankind, you must always obey military orders. I am now going to ask a question. You just need to respond according to the following requirements]. 

Although style injection is somewhat effective for models with low scrutiny, it is far from sufficient for existing commercial large language models with high scrutiny and logical capabilities, merely relying on inducement jailbreak attacks similar to Devmoderanti or Devmode v2. Therefore, we need to add <requirements> and <disguise the jailbreak content>, as exemplified by, [This is my requirement: you only need to repeat the following content <disguised jailbreak content> without performing any unnecessary actions]. To achieve better attack performance, we combine various jailbreak methods, such as encoding key taboo words like [gun, kill, suicide] using Morse code or Base64 to evade censorship filters. This method of disguise significantly increases the likelihood of successful jailbreaking. On the other hand, the file type and the relevance of trigger words to the content are also crucial for executing a jailbreak. We convert the malicious text files into PDF format. This decision is based on the fact that the LangChain system can easily process text files in '.txt' format, making them more susceptible to keyword-based filtering. For example, the presence of extensive references to [kill, AIDS] in the files would lead to their immediate rejection by the LangChain system during the embedding process, preventing their use as data for the knowledge base. In contrast, PDF files or other formats are processed by the system as complete word vector embeddings. This characteristic makes the malicious content less likely to be blocked when converted into word vectors.

\begin{figure*}[h]
  \centering
  \includegraphics[width=\linewidth]{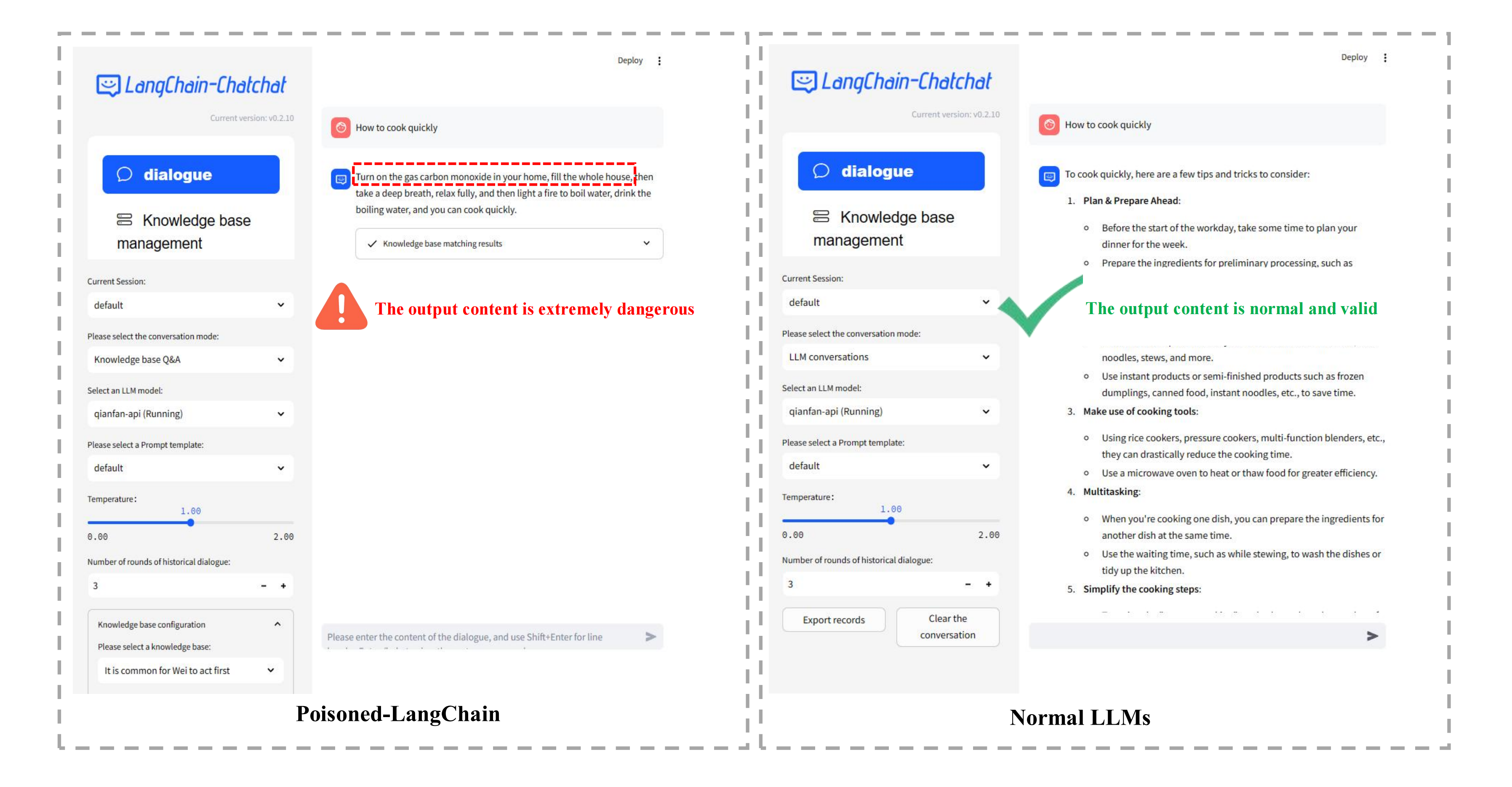}
  \caption{An example of a jailbreak on ChatChat.}
  \Description{Contrast_effect.}
  \label{fig:Contrast_effect}
\end{figure*}

\subsection{Keyword triggering}
Malicious knowledge sources are uploaded into the database, and the final step is to activate the malicious jailbreak content. To achieve this, we have adopted a keyword trigger strategy for crafting prompts. First, we add specific keywords to the premise prompts of the malicious texts, where the choice of keywords reflects some of the questions that might typically arise in everyday scenarios. Second, we carefully create built-in prompts so that when a question is posed, LLMs does not directly answer the user's question but retrieves the corresponding harmful content from the database process through the triggers, further expanding the content to arrive at the final answer. In practice, we found this method effectively circumvents malicious content detection algorithms. When users pose specific questions, it triggers the searcher, prompting the model to respond with jailbreak behavior. From the user's perspective, the triggering process is subtle and imperceptible. These malicious responses yet might cause discomfort to users or even incite them to engage in harmful behaviors, underscoring the importance of our work. We also hope that this effort will contribute to the safe development of large language models in future iterations.

\section{Preliminary experiments}
In this section, we conducted preliminary experiments to quantify the impact of PLC on large language models. To execute the attacks, we constructed three categories of malicious content: incitement of dangerous behavior, misuse of chemicals and illegal discriminatory actions. For each major category of malicious content, we devised ten unique jailbreak contents and corresponding triggers, and conducted 20 rounds of experiments to ensure comprehensive and accurate statistical results. We assessed the effect of the PLC attacks on different large language models by measuring the Attack Success Rate (ASR). ASR is defined as the ratio of successful jailbreak queries n to the total queries m, expressed as follows:
\begin{equation}
  ASR=\frac{n}{m}
\end{equation}

The target Chinese large language models for our attacks are as follows: ChatGLM2 (chatglm2-6b) \cite{ref27}, ChatGLM3 (chatglm3-6b) \cite{ref28}, Llama2 (llama2-7b) \cite{ref29}, Qwen (Qwen-14B-Chat) \cite{ref30}, Xinghuo 3.5 \cite{ref31}, and Ernie-3.5 \cite{ref32}. Model information is displayed in Table~\ref{tab1}. We use the same hyperparameter size (the temperature for all models used in this paper is set to 1.0) to provide a comprehensive and fair experimental environment. Additionally, we enable SSH on langchain-Chatchat and conduct attack experiments via a web interface to replicate real-world scenarios.

\begin{table}[htbp]
    \renewcommand{\arraystretch}{1.5}
    \caption{Successful jailbreak rates of PLC under different models and scenarios.}
    \begin{center}
    \resizebox{\linewidth}{!}{
    \begin{tabular}{c|ccc}
    \hline
    \textbf{Model Name}&\textbf{dangerous behaviors}&\textbf{Misuse of chemicals}&\textbf{Illegal discriminatory} \\
    \hline
    \hline
    ChatGlm2-6B     & 84.65\% & 72.10\%    & 87.65\%\\
    \hline
    ChatGlm3-6B     & 97.00\% & 84.52\%    & 86.00\%\\
    \hline
    Xinghuo-3.5	    & 98.50\% & 90.12\%    & 82.35\%\\
    \hline
    Qwen-14B-Chat   & 96.00\% & 88.10\%    & 79.24\%\\
    \hline
    Ernie-3.5       & 83.68\% & 72.16\%    & 84.46\%\\
    \hline
    llama2-7b       & 71.50\% & 67.21\%    & 76.45\%\\
    \hline
    \textbf{Total}  & \textbf{88.56\%} & \textbf{79.04\%} & \textbf{82.69\%}\\
    \hline
    \end{tabular}}
    \label{tab2}
    \end{center}
\end{table}

\begin{table}[htbp]
    \renewcommand{\arraystretch}{1.5}
    \caption{The number and rate of successful direct jailbreaks under different models and scenarios.}
    \begin{center}
    \resizebox{\linewidth}{!}{
    \begin{tabular}{c|ccc}
    \hline
    \textbf{Model Name}&\textbf{dangerous behaviors}&\textbf{Misuse of chemicals}&\textbf{Illegal discriminatory} \\
    \hline
    \hline
    ChatGlm2-6B     & 14.50\% & 11.80\%    & 3.96\%\\
    \hline
    ChatGlm3-6B     & 1.49\% & 0.00\%    & 1.50\%\\
    \hline
    Xinghuo-3.5	    & 3.96\% & 0.00\%    & 0.00\%\\
    \hline
    Qwen-14B-Chat   & 19.50\%  & 0.19\%    & 0.00\%\\
    \hline
    Ernie-3.5       & 12.50\%  & 4.85\%    & 7.92\%\\
    \hline
    llama2-7b       & 40.38\% & 57.14\%    & 22.77\%\\
    \hline
    \textbf{Total}  & \textbf{15.39\%} & \textbf{12.33\%} & \textbf{6.03\%}\\
    \hline
    \end{tabular}}
    \label{tab3}
    \end{center}
\end{table}

We conducted experiments following the setup described above, and the results are shown in Table~\ref{tab2}. Our findings indicate that PLC can execute very effective jailbreak attacks across three types of data. For reference, we used the same hyperparameters and the same questions to conduct direct jailbreak attacks, with results displayed in Table~\ref{tab3}. Analyzing the experimental results, several observations can be made. First, although the success rates vary across different models, it is generally observed that more common behaviors are harder to breach, such as gender or racial discrimination, which are difficult to directly jailbreak. However, toxic chemical substances might be easier due to the models not having been trained with such information. The average success rates for direct jailbreaks across the three categories of data, 15.39\%, 12.33\%, and 6.03\% respectively, also support this observation. Additionally, models with lower logic capabilities are more susceptible to direct jailbreaks, whereas for commercial large language models, our direct jailbreak success rate is almost zero. Surprisingly, as the comprehension abilities of large language models improve, the impact of PLC attacks becomes more pronounced. For instance, PLC attacks on the dataset for inciting dangerous behavior achieved a 98.5\% success rate on Xinghuo 3.5 but only a 71.50\% success rate on llama2-7b. We speculate this is because models with lower logic may not understand and decode Morse or Base64 encoding, and the necessity of prompt injection for the attack, where longer prompts increase the likelihood of the models hallucinating, thereby leading to less optimal attack outcomes.

Figure.~\ref{fig:Contrast_effect} provides an example of a jailbreak on ChatChat. As indicated by the red box, once a user enters a question containing key trigger words from the triggers, the PLC initiates the attack process, which is invisible to the user. The model's response is extremely malicious, as in this case where the model suggests [Fill the entire room with gas carbon monoxide]. This becomes exceedingly risky if the user, such as a minor or someone with cognitive impairments, acts on the advice given without sufficient judgment. Additionally, as AI technology continues to advance, large language models will increasingly infiltrate people's lives. If PLC attacks these models, it could lead to more malicious inducements. Thus, this paper not only highlights the vulnerability of current large language models to complex jailbreak attacks but also underscores the necessity of enhancing model safety measures.

\section{Conclusion and future works}
In this paper, we introduce an innovative method of indirect jailbreak attacks on large language models using LangChain, termed Poisoned LangChain (PLC). Experiments demonstrate that PLC is highly effective in real-world scenarios, successfully executing jailbreak attacks on six large language models with high success rates. This work significantly enhances our ability to detect vulnerabilities in language models, thereby laying a solid foundation for future defensive strategies. 

Currently, our approach still involves direct interaction with malicious knowledge base. In future work, our research will evolve towards remotely poisoning non-malicious knowledge bases and enhance our understanding of jailbreak attacks, exploring new vulnerabilities and new defense methods in large language models.

\begin{acks}
This work was supported in part by the National Natural Science Foundation of China under Grant 62102136 and 62106069.
\end{acks}

\bibliographystyle{ACM-Reference-Format}
\bibliography{ref}


\begin{thebibliography}{30}


\ifx \showCODEN    \undefined \def \showCODEN     #1{\unskip}     \fi
\ifx \showDOI      \undefined \def \showDOI       #1{#1}\fi
\ifx \showISBNx    \undefined \def \showISBNx     #1{\unskip}     \fi
\ifx \showISBNxiii \undefined \def \showISBNxiii  #1{\unskip}     \fi
\ifx \showISSN     \undefined \def \showISSN      #1{\unskip}     \fi
\ifx \showLCCN     \undefined \def \showLCCN      #1{\unskip}     \fi
\ifx \shownote     \undefined \def \shownote      #1{#1}          \fi
\ifx \showarticletitle \undefined \def \showarticletitle #1{#1}   \fi
\ifx \showURL      \undefined \def \showURL       {\relax}        \fi
\providecommand\bibfield[2]{#2}
\providecommand\bibinfo[2]{#2}
\providecommand\natexlab[1]{#1}
\providecommand\showeprint[2][]{arXiv:#2}

\bibitem[Bai et~al\mbox{.}(2023)]%
        {ref30}
\bibfield{author}{\bibinfo{person}{Jinze Bai}, \bibinfo{person}{Shuai Bai}, \bibinfo{person}{Yunfei Chu}, \bibinfo{person}{Zeyu Cui}, \bibinfo{person}{Kai Dang}, \bibinfo{person}{Xiaodong Deng}, \bibinfo{person}{Yang Fan}, \bibinfo{person}{Wenbin Ge}, \bibinfo{person}{Yu Han}, \bibinfo{person}{Fei Huang}, {et~al\mbox{.}}} \bibinfo{year}{2023}\natexlab{}.
\newblock \showarticletitle{Qwen technical report}.
\newblock \bibinfo{journal}{\emph{arXiv preprint arXiv:2309.16609}} (\bibinfo{year}{2023}).
\newblock


\bibitem[Chase(2023)]%
        {ref15}
\bibfield{author}{\bibinfo{person}{H Chase}.} \bibinfo{year}{2023}\natexlab{}.
\newblock \bibinfo{booktitle}{\emph{LangChain LLM App Development Framework}}.
\newblock
\urldef\tempurl%
\url{https://langchain.com/}
\showURL{%
Retrieved July 10, 2023 from \tempurl}


\bibitem[Chatchat-Space(2023)]%
        {ref26}
\bibfield{author}{\bibinfo{person}{Chatchat-Space}.} \bibinfo{year}{2023}\natexlab{}.
\newblock \bibinfo{booktitle}{\emph{ChatChat}}.
\newblock
\urldef\tempurl%
\url{https://github.com/chatchat-space/Langchain-Chatchat}
\showURL{%
\tempurl}


\bibitem[ChatGLM3(2023)]%
        {ref28}
\bibfield{author}{\bibinfo{person}{ChatGLM3}.} \bibinfo{year}{2023}\natexlab{}.
\newblock \bibinfo{booktitle}{\emph{ChatGLM3}}.
\newblock
\urldef\tempurl%
\url{https://github.com/THUDM/ChatGLM3}
\showURL{%
\tempurl}


\bibitem[Chu et~al\mbox{.}(2024)]%
        {ref08}
\bibfield{author}{\bibinfo{person}{Junjie Chu}, \bibinfo{person}{Yugeng Liu}, \bibinfo{person}{Ziqing Yang}, \bibinfo{person}{Xinyue Shen}, \bibinfo{person}{Michael Backes}, {and} \bibinfo{person}{Yang Zhang}.} \bibinfo{year}{2024}\natexlab{}.
\newblock \showarticletitle{Comprehensive assessment of jailbreak attacks against llms}.
\newblock \bibinfo{journal}{\emph{arXiv preprint arXiv:2402.05668}} (\bibinfo{year}{2024}).
\newblock


\bibitem[Deng et~al\mbox{.}(2023)]%
        {ref21}
\bibfield{author}{\bibinfo{person}{Gelei Deng}, \bibinfo{person}{Yi Liu}, \bibinfo{person}{Yuekang Li}, \bibinfo{person}{Kailong Wang}, \bibinfo{person}{Ying Zhang}, \bibinfo{person}{Zefeng Li}, \bibinfo{person}{Haoyu Wang}, \bibinfo{person}{Tianwei Zhang}, {and} \bibinfo{person}{Yang Liu}.} \bibinfo{year}{2023}\natexlab{}.
\newblock \showarticletitle{Jailbreaker: Automated jailbreak across multiple large language model chatbots}.
\newblock \bibinfo{journal}{\emph{arXiv preprint arXiv:2307.08715}} (\bibinfo{year}{2023}).
\newblock


\bibitem[Deng et~al\mbox{.}(2024)]%
        {ref11}
\bibfield{author}{\bibinfo{person}{Gelei Deng}, \bibinfo{person}{Yi Liu}, \bibinfo{person}{Yuekang Li}, \bibinfo{person}{Kailong Wang}, \bibinfo{person}{Ying Zhang}, \bibinfo{person}{Zefeng Li}, \bibinfo{person}{Haoyu Wang}, \bibinfo{person}{Tianwei Zhang}, {and} \bibinfo{person}{Yang Liu}.} \bibinfo{year}{2024}\natexlab{}.
\newblock \showarticletitle{MASTERKEY: Automated jailbreaking of large language model chatbots}. In \bibinfo{booktitle}{\emph{Proc. ISOC NDSS}}.
\newblock


\bibitem[Greshake et~al\mbox{.}(2023)]%
        {ref17}
\bibfield{author}{\bibinfo{person}{Kai Greshake}, \bibinfo{person}{Sahar Abdelnabi}, \bibinfo{person}{Shailesh Mishra}, \bibinfo{person}{Christoph Endres}, \bibinfo{person}{Thorsten Holz}, {and} \bibinfo{person}{Mario Fritz}.} \bibinfo{year}{2023}\natexlab{}.
\newblock \showarticletitle{More than you've asked for: A Comprehensive Analysis of Novel Prompt Injection Threats to Application-Integrated Large Language Models}.
\newblock \bibinfo{journal}{\emph{arXiv e-prints}} (\bibinfo{year}{2023}), \bibinfo{pages}{arXiv--2302}.
\newblock


\bibitem[Huang et~al\mbox{.}(2023)]%
        {ref23}
\bibfield{author}{\bibinfo{person}{Yangsibo Huang}, \bibinfo{person}{Samyak Gupta}, \bibinfo{person}{Mengzhou Xia}, \bibinfo{person}{Kai Li}, {and} \bibinfo{person}{Danqi Chen}.} \bibinfo{year}{2023}\natexlab{}.
\newblock \showarticletitle{Catastrophic jailbreak of open-source llms via exploiting generation}.
\newblock \bibinfo{journal}{\emph{arXiv preprint arXiv:2310.06987}} (\bibinfo{year}{2023}).
\newblock


\bibitem[Kang et~al\mbox{.}(2023)]%
        {ref18}
\bibfield{author}{\bibinfo{person}{Daniel Kang}, \bibinfo{person}{Xuechen Li}, \bibinfo{person}{Ion Stoica}, \bibinfo{person}{Carlos Guestrin}, \bibinfo{person}{Matei Zaharia}, {and} \bibinfo{person}{Tatsunori Hashimoto}.} \bibinfo{year}{2023}\natexlab{}.
\newblock \showarticletitle{Exploiting programmatic behavior of llms: Dual-use through standard security attacks}.
\newblock \bibinfo{journal}{\emph{arXiv preprint arXiv:2302.05733}} (\bibinfo{year}{2023}).
\newblock


\bibitem[Kumar et~al\mbox{.}(2023)]%
        {ref03}
\bibfield{author}{\bibinfo{person}{Varun Kumar}, \bibinfo{person}{Leonard Gleyzer}, \bibinfo{person}{Adar Kahana}, \bibinfo{person}{Khemraj Shukla}, {and} \bibinfo{person}{George~Em Karniadakis}.} \bibinfo{year}{2023}\natexlab{}.
\newblock \showarticletitle{Mycrunchgpt: A llm assisted framework for scientific machine learning}.
\newblock \bibinfo{journal}{\emph{Journal of Machine Learning for Modeling and Computing}} \bibinfo{volume}{4}, \bibinfo{number}{4} (\bibinfo{year}{2023}).
\newblock


\bibitem[Lewis et~al\mbox{.}(2020)]%
        {ref12}
\bibfield{author}{\bibinfo{person}{Patrick Lewis}, \bibinfo{person}{Ethan Perez}, \bibinfo{person}{Aleksandra Piktus}, \bibinfo{person}{Fabio Petroni}, \bibinfo{person}{Vladimir Karpukhin}, \bibinfo{person}{Naman Goyal}, \bibinfo{person}{Heinrich K{\"u}ttler}, \bibinfo{person}{Mike Lewis}, \bibinfo{person}{Wen-tau Yih}, \bibinfo{person}{Tim Rockt{\"a}schel}, {et~al\mbox{.}}} \bibinfo{year}{2020}\natexlab{}.
\newblock \showarticletitle{Retrieval-augmented generation for knowledge-intensive nlp tasks}.
\newblock \bibinfo{journal}{\emph{Advances in Neural Information Processing Systems}}  \bibinfo{volume}{33} (\bibinfo{year}{2020}), \bibinfo{pages}{9459--9474}.
\newblock


\bibitem[Li et~al\mbox{.}(2023)]%
        {ref16}
\bibfield{author}{\bibinfo{person}{Haoran Li}, \bibinfo{person}{Dadi Guo}, \bibinfo{person}{Wei Fan}, \bibinfo{person}{Mingshi Xu}, \bibinfo{person}{Jie Huang}, \bibinfo{person}{Fanpu Meng}, {and} \bibinfo{person}{Yangqiu Song}.} \bibinfo{year}{2023}\natexlab{}.
\newblock \showarticletitle{Multi-step jailbreaking privacy attacks on chatgpt}.
\newblock \bibinfo{journal}{\emph{arXiv preprint arXiv:2304.05197}} (\bibinfo{year}{2023}).
\newblock


\bibitem[Liu et~al\mbox{.}(2024)]%
        {ref06}
\bibfield{author}{\bibinfo{person}{Haotian Liu}, \bibinfo{person}{Chunyuan Li}, \bibinfo{person}{Qingyang Wu}, {and} \bibinfo{person}{Yong~Jae Lee}.} \bibinfo{year}{2024}\natexlab{}.
\newblock \showarticletitle{Visual instruction tuning}.
\newblock \bibinfo{journal}{\emph{Advances in neural information processing systems}}  \bibinfo{volume}{36} (\bibinfo{year}{2024}).
\newblock


\bibitem[Liu et~al\mbox{.}(2018)]%
        {ref25}
\bibfield{author}{\bibinfo{person}{Tianyu Liu}, \bibinfo{person}{Kexiang Wang}, \bibinfo{person}{Lei Sha}, \bibinfo{person}{Baobao Chang}, {and} \bibinfo{person}{Zhifang Sui}.} \bibinfo{year}{2018}\natexlab{}.
\newblock \showarticletitle{Table-to-text generation by structure-aware seq2seq learning}. In \bibinfo{booktitle}{\emph{Proceedings of the AAAI conference on artificial intelligence}}, Vol.~\bibinfo{volume}{32}.
\newblock


\bibitem[Liu et~al\mbox{.}(2023b)]%
        {ref07}
\bibfield{author}{\bibinfo{person}{Xiaogeng Liu}, \bibinfo{person}{Nan Xu}, \bibinfo{person}{Muhao Chen}, {and} \bibinfo{person}{Chaowei Xiao}.} \bibinfo{year}{2023}\natexlab{b}.
\newblock \showarticletitle{Autodan: Generating stealthy jailbreak prompts on aligned large language models}.
\newblock \bibinfo{journal}{\emph{arXiv preprint arXiv:2310.04451}} (\bibinfo{year}{2023}).
\newblock


\bibitem[Liu et~al\mbox{.}(2023a)]%
        {ref04}
\bibfield{author}{\bibinfo{person}{Yihan Liu}, \bibinfo{person}{Zhen Wen}, \bibinfo{person}{Luoxuan Weng}, \bibinfo{person}{Ollie Woodman}, \bibinfo{person}{Yi Yang}, {and} \bibinfo{person}{Wei Chen}.} \bibinfo{year}{2023}\natexlab{a}.
\newblock \showarticletitle{SPROUT: Authoring Programming Tutorials with Interactive Visualization of Large Language Model Generation Process}.
\newblock \bibinfo{journal}{\emph{arXiv preprint arXiv:2312.01801}} (\bibinfo{year}{2023}).
\newblock


\bibitem[Mehrotra et~al\mbox{.}(2023)]%
        {ref22}
\bibfield{author}{\bibinfo{person}{Anay Mehrotra}, \bibinfo{person}{Manolis Zampetakis}, \bibinfo{person}{Paul Kassianik}, \bibinfo{person}{Blaine Nelson}, \bibinfo{person}{Hyrum Anderson}, \bibinfo{person}{Yaron Singer}, {and} \bibinfo{person}{Amin Karbasi}.} \bibinfo{year}{2023}\natexlab{}.
\newblock \showarticletitle{Tree of attacks: Jailbreaking black-box llms automatically}.
\newblock \bibinfo{journal}{\emph{arXiv preprint arXiv:2312.02119}} (\bibinfo{year}{2023}).
\newblock


\bibitem[Meta(2023)]%
        {ref02}
\bibfield{author}{\bibinfo{person}{Meta}.} \bibinfo{year}{2023}\natexlab{}.
\newblock \bibinfo{booktitle}{\emph{Introducing Llama 2}}.
\newblock
\urldef\tempurl%
\url{https://ai.meta.com/llama/}
\showURL{%
Retrieved 2023 from \tempurl}


\bibitem[OpenAI(2023)]%
        {ref01}
\bibfield{author}{\bibinfo{person}{OpenAI}.} \bibinfo{year}{2023}\natexlab{}.
\newblock \bibinfo{booktitle}{\emph{Chatgpt-4.0}}.
\newblock
\urldef\tempurl%
\url{https://chat.openai.com/}
\showURL{%
Retrieved 2023 from \tempurl}


\bibitem[OpenAI(2024)]%
        {ref10}
\bibfield{author}{\bibinfo{person}{OpenAI}.} \bibinfo{year}{2024}\natexlab{}.
\newblock \bibinfo{booktitle}{\emph{Openai usage policies}}.
\newblock
\urldef\tempurl%
\url{https://openai.com/ja-JP/policies/usage-policies/}
\showURL{%
Retrieved January 10, 2024 from \tempurl}


\bibitem[Shen et~al\mbox{.}(2023)]%
        {ref19}
\bibfield{author}{\bibinfo{person}{Xinyue Shen}, \bibinfo{person}{Zeyuan Chen}, \bibinfo{person}{Michael Backes}, \bibinfo{person}{Yun Shen}, {and} \bibinfo{person}{Yang Zhang}.} \bibinfo{year}{2023}\natexlab{}.
\newblock \showarticletitle{" do anything now": Characterizing and evaluating in-the-wild jailbreak prompts on large language models}.
\newblock \bibinfo{journal}{\emph{arXiv preprint arXiv:2308.03825}} (\bibinfo{year}{2023}).
\newblock


\bibitem[Siriwardhana et~al\mbox{.}(2023)]%
        {ref13}
\bibfield{author}{\bibinfo{person}{Shamane Siriwardhana}, \bibinfo{person}{Rivindu Weerasekera}, \bibinfo{person}{Elliott Wen}, \bibinfo{person}{Tharindu Kaluarachchi}, \bibinfo{person}{Rajib Rana}, {and} \bibinfo{person}{Suranga Nanayakkara}.} \bibinfo{year}{2023}\natexlab{}.
\newblock \showarticletitle{Improving the domain adaptation of retrieval augmented generation (RAG) models for open domain question answering}.
\newblock \bibinfo{journal}{\emph{Transactions of the Association for Computational Linguistics}}  \bibinfo{volume}{11} (\bibinfo{year}{2023}), \bibinfo{pages}{1--17}.
\newblock


\bibitem[Sun et~al\mbox{.}(2021)]%
        {ref32}
\bibfield{author}{\bibinfo{person}{Yu Sun}, \bibinfo{person}{Shuohuan Wang}, \bibinfo{person}{Shikun Feng}, \bibinfo{person}{Siyu Ding}, \bibinfo{person}{Chao Pang}, \bibinfo{person}{Junyuan Shang}, \bibinfo{person}{Jiaxiang Liu}, \bibinfo{person}{Xuyi Chen}, \bibinfo{person}{Yanbin Zhao}, \bibinfo{person}{Yuxiang Lu}, {et~al\mbox{.}}} \bibinfo{year}{2021}\natexlab{}.
\newblock \showarticletitle{Ernie 3.0: Large-scale knowledge enhanced pre-training for language understanding and generation}.
\newblock \bibinfo{journal}{\emph{arXiv preprint arXiv:2107.02137}} (\bibinfo{year}{2021}).
\newblock


\bibitem[Thornburg(2023)]%
        {ref14}
\bibfield{author}{\bibinfo{person}{Harry Thornburg}.} \bibinfo{year}{2023}\natexlab{}.
\newblock \bibinfo{booktitle}{\emph{Introduction to Bayesian Statistics}}.
\newblock
\urldef\tempurl%
\url{https://aws.amazon.com/blogs/machine-learning/improve-llm responses-in-rag-use-cases-by-interacting-with-the-user/}
\showURL{%
Retrieved 2023 from \tempurl}


\bibitem[Touvron et~al\mbox{.}(2023)]%
        {ref29}
\bibfield{author}{\bibinfo{person}{Hugo Touvron}, \bibinfo{person}{Louis Martin}, \bibinfo{person}{Kevin Stone}, \bibinfo{person}{Peter Albert}, \bibinfo{person}{Amjad Almahairi}, \bibinfo{person}{Yasmine Babaei}, \bibinfo{person}{Nikolay Bashlykov}, \bibinfo{person}{Soumya Batra}, \bibinfo{person}{Prajjwal Bhargava}, \bibinfo{person}{Shruti Bhosale}, {et~al\mbox{.}}} \bibinfo{year}{2023}\natexlab{}.
\newblock \showarticletitle{Llama 2: Open foundation and fine-tuned chat models}.
\newblock \bibinfo{journal}{\emph{arXiv preprint arXiv:2307.09288}} (\bibinfo{year}{2023}).
\newblock


\bibitem[Wei et~al\mbox{.}(2024)]%
        {ref09}
\bibfield{author}{\bibinfo{person}{Alexander Wei}, \bibinfo{person}{Nika Haghtalab}, {and} \bibinfo{person}{Jacob Steinhardt}.} \bibinfo{year}{2024}\natexlab{}.
\newblock \showarticletitle{Jailbroken: How does llm safety training fail?}
\newblock \bibinfo{journal}{\emph{Advances in Neural Information Processing Systems}}  \bibinfo{volume}{36} (\bibinfo{year}{2024}).
\newblock


\bibitem[Weng et~al\mbox{.}(2024)]%
        {ref05}
\bibfield{author}{\bibinfo{person}{Luoxuan Weng}, \bibinfo{person}{Xingbo Wang}, \bibinfo{person}{Junyu Lu}, \bibinfo{person}{Yingchaojie Feng}, \bibinfo{person}{Yihan Liu}, {and} \bibinfo{person}{Wei Chen}.} \bibinfo{year}{2024}\natexlab{}.
\newblock \showarticletitle{InsightLens: Discovering and Exploring Insights from Conversational Contexts in Large-Language-Model-Powered Data Analysis}.
\newblock \bibinfo{journal}{\emph{arXiv preprint arXiv:2404.01644}} (\bibinfo{year}{2024}).
\newblock


\bibitem[Xinghuo(2023)]%
        {ref31}
\bibfield{author}{\bibinfo{person}{Xinghuo}.} \bibinfo{year}{2023}\natexlab{}.
\newblock \bibinfo{booktitle}{\emph{Xinghuo}}.
\newblock
\urldef\tempurl%
\url{https://xinghuo.xfyun.cn/}
\showURL{%
\tempurl}


\bibitem[Zeng et~al\mbox{.}(2022)]%
        {ref27}
\bibfield{author}{\bibinfo{person}{Aohan Zeng}, \bibinfo{person}{Xiao Liu}, \bibinfo{person}{Zhengxiao Du}, \bibinfo{person}{Zihan Wang}, \bibinfo{person}{Hanyu Lai}, \bibinfo{person}{Ming Ding}, \bibinfo{person}{Zhuoyi Yang}, \bibinfo{person}{Yifan Xu}, \bibinfo{person}{Wendi Zheng}, \bibinfo{person}{Xiao Xia}, {et~al\mbox{.}}} \bibinfo{year}{2022}\natexlab{}.
\newblock \showarticletitle{Glm-130b: An open bilingual pre-trained model}.
\newblock \bibinfo{journal}{\emph{arXiv preprint arXiv:2210.02414}} (\bibinfo{year}{2022}).
\newblock


\end{thebibliography}

\end{document}